%% file: root.tex
\documentclass[letterpaper, 10 pt, conference]{ieeeconf}  

\IEEEoverridecommandlockouts                              

\pdfoutput=1
\UseRawInputEncoding

\usepackage{amsmath} 
\usepackage{amsfonts}
\usepackage{amssymb}
\usepackage{algorithm}
\usepackage{algorithmic}

\usepackage{bm}
\usepackage{cite}
\usepackage{epsfig}
\usepackage{float}
\usepackage{flushend}
\usepackage{graphics} 
\usepackage{graphicx}
\usepackage[disallowspaces]{mathtools}
\usepackage{mathptmx} 
\usepackage{multicol}
\usepackage{paralist}
\usepackage{subfigure}
\usepackage{times}
\usepackage{tikz}
\usetikzlibrary{shapes,snakes,automata,backgrounds}

\newcommand{\EQ}{\begin{eqnarray}}
\newcommand{\EN}{\end{eqnarray}}
\newcommand{\EQQ}{\begin{eqnarray*}}
\newcommand{\ENN}{\end{eqnarray*}}
\newcommand{\BM}{\begin{bmatrix}}
\newcommand{\EM}{\end{bmatrix}}
\newcommand{\PBM}{\begin{pmatrix}}
\newcommand{\PEM}{\end{pmatrix}}
\newcommand{\BC}{\begin{cases}}
\newcommand{\EC}{\end{cases}}
\newcommand{\MB}{\begin{bmatrix}}
\newcommand{\MBN}{\end{bmatrix}}
\newcommand{\MBM}{\big[\begin{matrix}}
\newcommand{\MBNM}{\end{matrix}\big]}
\newdimen\figrasterwd
\figrasterwd\textwidth
\input{articles/title_ieee.tex}

\begin{document}

\maketitle
\thispagestyle{empty}
\pagestyle{empty}

\input{articles/abstract.tex}
\input{articles/introduction.tex}

\input{articles/related.tex}

\input{articles/preliminary.tex}

\input{articles/methodology.tex}
\input{articles/experiment.tex}



\end{document}

%% file: articles/title_ieee.tex
\title{\LARGE \bf
Incremental Semantic Localization using\\Hierarchical Clustering of Object Association Sets}

\author{Lan Hu$^{\times,*}$, Zhongwei Luo$^{\times}$, Runze Yuan$^{\times}$, Yuchen  Cao$^{\times}$, Jiaxin Wei$^{\times}$, \\ Kai Wang$^{+}$ and Laurent Kneip$^{\times}$
\thanks{$^{\times}$Mobile Perception Lab, SIST, ShanghaiTech University.
\newline
        {\tt\small \{hulan,luozhw, yuanrz\}@shanghaitech.edu.cn}
        {\tt\small\{caoych,weijx,lkneip\}@shanghaitech.edu.cn}
        }%
\thanks{$^{*}$Shanghai Institute of Microsystem and  Information Technology, Chinese Academy of Sciences and University of Chinese Academy of Sciences. }

\thanks{$^{+}$Application Lab, Huawei Incorporated Company, P.R. China. \newline
{\tt\small\{wangkai197\}@huawei.com}}
}

%% file: articles/abstract.tex
\begin{abstract}
We present a novel approach for relocalization or place recognition, a fundamental problem to be solved in many robotics, automation, and AR applications. Rather than relying on often unstable appearance information, we consider a situation in which the reference map is given in the form of localized objects. Our localization framework relies on 3D semantic object detections, which are then associated to objects in the map. Possible pair-wise association sets are grown based on hierarchical clustering using a merge metric that evaluates spatial compatibility. The latter notably uses information about relative object configurations, which is invariant with respect to global transformations. Association sets are furthermore updated and expanded as the camera incrementally explores the environment and detects further objects. We test our algorithm in several challenging situations including dynamic scenes, large view-point changes, and scenes with repeated instances. Our experiments demonstrate that our approach outperforms prior art in terms of both robustness and accuracy.
\end{abstract}

%% file: articles/introduction.tex

\section{Introduction}
\label{sec:introduction}
Visual localization estimates the 6 Degree-of-Freedom (DoF) pose of a camera view observing a 3D scene. It plays an essential role in a wide range of practical applications, such as robotics, augmented reality, and smart vehicles. One of the major challenges behind localization revolves around the reliable recognition of places under both a wide variety of environmental conditions that non-uniformly affect the appearance of the scene, as well as a wide range of viewpoints. For long-term outdoor autonomous navigation, there are many disturbing factors such as seasonal changes or weather conditions, natural or artificial illumination variations, and varying positions of for example cars and persons. To deal with such challenges, the stability of traditional, hand-crafted low-level features is generally deemed insufficient.

With the development of deep neural networks, there nowadays exist several novel algorithms able to produce stable higher-level features \cite{wu2016learning} or directly predict the camera pose \cite{naseer2017deep}. Many such algorithms empower platforms with a strong basis for long-term navigation and 3D scene understanding, and support operation in both indoor and outdoor environments. Compared against outdoor localization, the indoor localization problem is significantly easier owing to the possibility of using a depth camera and thus further reduce the influence of appearance changes. However, benefits in both environments may be obtained from object detection \cite{ren2015faster,redmon2016you} or instance segmentation networks \cite{chen2020blendmask,bolya2019yolact,johnson2018adapting}, both of which produce rich and discriminative semantic information about the environment. Compared against low-level point features, object landmark detections are conceived to be much more stable under a variety of illumination and viewpoint changes. 

Human beings are excellent at place recognition and localization by utilizing semantic landmarks and their relationships. For example, given a view of a corridor and the respective arrangement of doors, we may easily know in which part of a building we are located.
The more landmarks are observed, the more confidently and accurately we can locate ourselves. Our framework employs a similar strategy and aims at localization by identifying a subset of objects in a map with a semantically and spatially consistent pair-wise association to a locally measured set of objects. The confidence of a pair-wise association set depends on the similarity of the assigned object sets expressed in terms of stable properties such as appearance-independent features and relative spatial object arrangement. The considered information is invariant with respect to viewpoint and illumination variations. The approach works incrementally and expands the set of hypothesized association sets as the camera explores the environment and detects new objects. Association sets are furthermore grown based on hierarchical clustering using a merge metric that depends on the said spatial compatibilities. Once a dominant pair-wise association set has been confirmed, a subsequent alignment of the assigned objects reveals the absolute camera position and orientation.

Our contributions are summarized as follows:
 \begin{itemize}
	\item A light-weight, semantic landmark based relocalization solution for indoor scenes.
	\item The method constructs pair-wise association sets between local object measurements and a prior given object-level map based on hierarchical clustering. Confidence scores are given by considering consistency in terms of semantics and spatial relative arrangements.
 	\item The method works incrementally and expands the association sets as more objects are being detected.
 	\item A diverse set of experiments verifies the effectiveness of our proposed solution as well as an obvious improvement over the previous state-of-the-art.
 \end{itemize}

%% file: articles/related.tex

\section{Related Work}
\label{sec:related}
\textbf{Point-level features for relocalization:}
The most common relocalization or place recognition methods in SLAM exploit local invariant key-points such as SIFT~\cite{ng2003sift} or ORB~\cite{mur2015orb}. The most straightforward solution is given by a direct establishment of 2D-3D matches between a query image and 3D points in a map, followed by the execution of a robust camera resectioning algorithm for geometric verification~\cite{fischler1981random}. Alternatively, the scalability of the place recognition module may be increased by adding the bag-of-words approach \cite{arandjelovic2014visual}, likely in conjunction with an inverted index for efficient image retrieval~\cite{nister2006scalable}. However, the approach remains highly sensitive to illumination or even camera view-point changes, which may impact significantly on the appearance of low-level features. In an aim to provide improved invariance over hand-designed features, more recent works have proposed learned 2D feature point extractors \cite{simonyan2014learning,yi2016lift,detone2018superpoint}. However, Jin et al.~\cite{jin2020image} point out that the performance of traditional features such as SIFT is still much more robust and stable than learned features in many practical situations. If available, depth information can be used to extract 3D descriptors~\cite{khoury2017learning,rusu2009fast,zeng20173dmatch,gojcic2019perfect,junyi2021camera}. Sometimes, depth-supported approaches still fail in the presence of strong viewpoint or appearance changes if lacking visual or structural overlap.

Another popular strategy for point-based localization \cite{kobyshev2014matching,mousavian2015semantically,toft2018semantic} leverages semantic information to re-weight or discard ambiguous features \cite{knopp2010avoiding} and increase the quality of the 2D-3D correspondences. Similarly, bag-of-words representations can be enhanced by combining semantic local features~\cite{arandjelovic2014visual}. While there are many ways of including semantics as auxiliary information, they will not change the fact that point-level features remain highly sensitive to illumination or view-point changes.

\textbf{High-level features for relocalization:}
Rather than using the features of key-points, the community has proposed several alternative directions for place recognition such as learning tailored representations of query data  \cite{mousavian2015semantically,ulrich2000appearance,angelina2018pointnetvlad,chen2014convolutional,arandjelovic2016netvlad,schonberger2018semantic,taira2018inloc} or directly regressing the camera pose of the query data by a neural network \cite{brachmann2017dsac,walch2017image,parisotto2018global,lin2019deep,naseer2017deep,ding2019camnet}. 
The main drawback of these approaches is that they need to be retrained for each dataset in order to get satisfactory and stable results. Furthermore, these methods have insufficient robustness for stand-alone operation as they depend on post-processing steps to verify geometric or photometric consistency. Their preferred application scenario is therefore given by outdoor long-term navigation with low accuracy requirements. Indoor localization, on the other hand, requires much more accurate estimation and may additionally suffer from challenges such as differing sensors for the construction of the global map prior and the online measurements.

Segmatch~\cite{dube2016segmatch} adopts an object feature extraction and matching strategy to find corresponding objects between the local and the global map. The relocalization is finalized by calculating the camera-to-model pose from the identified object correspondences. However, object shape differences caused by camera viewpoint variations easily impact on the stability of the detected features, and thus the overall relocalization quality.
Further works that treat objects as landmarks for relocalization are given by \cite{li2019semantic} and \cite{li2018semantic}, who rely on 3D semantic cube consistency. There are several more examples of works relying on semantic information  \cite{dube2016segmatch,speciale2018consensus,gaudilliere2019camera,zhang2019hierarchical}. However, as for previous examples, they fail to exploit object-to-object relationships for relocalization.

The most related work to ours is given by Liu et al.~\cite{liu2019global}, who pose the relocalization problem as a graph matching problem in which objects are taken as nodes to build a graph topology~\cite{gawel2018x}. They use a random walk method to record a node's semantic information as a descriptor for object association. By counting the number of matched descriptors, they choose the top-k objects from the global map and set them as potential correspondences. While highly related, the purely semantic descriptor makes the algorithm easily fail if the environment contains several repeated instances of semantically comparable object clusters. In contrast to our method, the algorithm depends on a predefined walk length which makes the algorithm unsuitable for the incremental relocalization. It might fail with high probability in the situations which contain multiple same category instances since that the predefined walk length will limit the ability to 
accurately describe the object spatial distribution.

%% file: articles/preliminary.tex

\section{Preliminaries}
\label{sec:preliminary}
\textbf{Notations and problem definition:}
The global semantic map is represented by the set $ \mathbf{A} = \{ \mathbf{A}_i \}_{i=1, \cdots, N_A}$, where $\mathbf{A}_i = \{ \mathbf{P}^A_i, s^A_i, \mathbf{m}^A_i\}$ are object information bags and $N_{A}$ is the number of objects. $\mathbf{A}_i$ contains an object point set $\mathbf{P}^A_i$, semantic label $s^A_i$, and  object position $\mathbf{m}^A_i$. Similarly, $\mathbf{B}=\{ \mathbf{B}_i\}_{i=1, \cdots, N_{B}}$ contains the objects in the local map. Let $\mathbf{D} = \{\mathbf{D}_k|k=1,\cdots, K \}$ denote the set of all hypothesized object association sets. $K$ is the number of association sets which has a predefined upper bound to limit computational load.
Each $\mathbf{D}_k = \{(\alpha_i,\beta_i)|_{i=1, \cdots, |\mathbf{D}_k|} ,v_k\}$ contains a set of pair-wise correspondences indicating associations between the $\alpha_i$-th global object and the $\beta_i$-th local object. $|\mathbf{D}_k|$ is the number of pair-wise object correspondences, and $v_k$ indicates the score (or vote) for a pair-wise association set.
The objective of our object-based localization is given by estimating the transformation that aligns the current local observation with objects in the global map, that is 
\begin{equation}
\mathbf{x}^*= \arg \max_{\mathbf{x}}  p( \mathbf{B}, \mathbf{D}|\mathbf{A},  \mathbf{x}),
\end{equation}
where $\mathbf{x}$ is the  camera pose parameter.
It is difficult to solve this problem directly, as the correspondences are a latent variable for which no prior information is given. 
General techniques for semantic relocalization are therefore given by first finding the association set $\mathbf{D}$ (i.e. the correspondences), followed by estimating the camera pose.

\textbf{Hierarchical clustering:} Also known as hierarchical cluster analysis, the algorithm groups compatible samples into clusters. Unlike k-means and GMM clustering, it does not require prior knowledge about the number of clusters. Strategies for hierarchical clustering generally fall into two categories: \textit{agglomerative} and \textit{divisive}. Here we only use the agglomerative strategy.
It starts by treating each sample as a separate cluster. Then, it repeatedly executes the following two steps: (1) identify two clusters that can be merged, and (2) conditionally merge the two clusters if a merge criterion is met. The iterative process continues until no more clusters can be merged.
In most hierarchical clustering methods, compatibility is expressed by a metric such as the distance between pairs of samples. The linkage criterion then decides if clusters should be merged based the value of the compatibility metric.
%
For more information about hierarchical clustering, please refer to \cite{maimon2005data}. Note that in our case, clusters are given by association sets, and samples are given by individual pair-wise associations.

%% file: articles/methodology.tex
\section{Methodology}
\label{sec:methodology}
\begin{figure*}[htpb]
	\centering
	\includegraphics[scale=0.85]{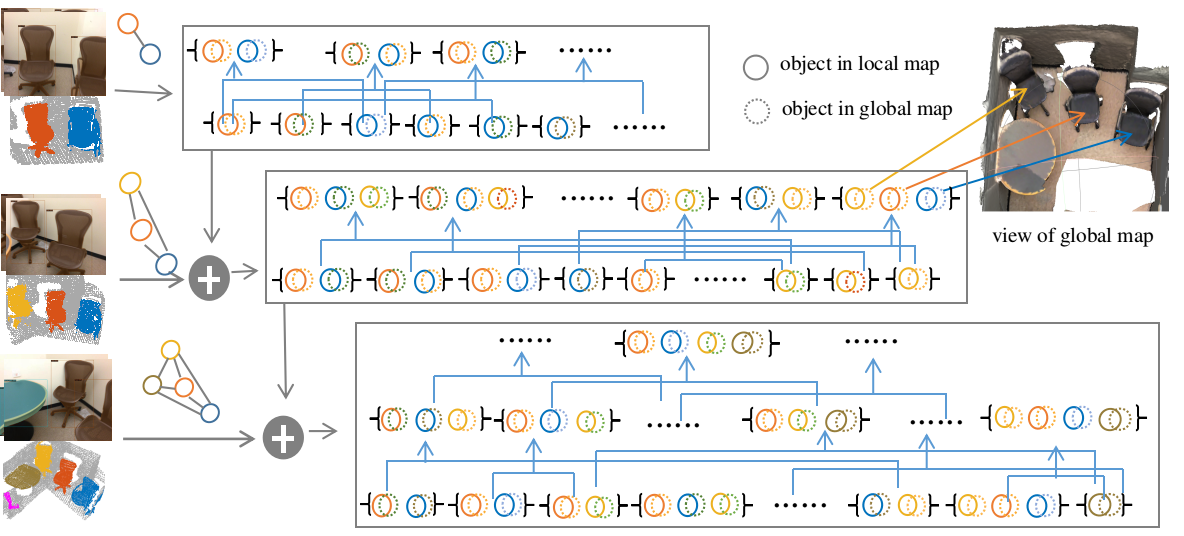}
	\caption{The Pipeline of the semantic incremental localization. We build the instance-level map and construct the fully connected graph. For each new observation, we first find its potential correspondences in the global, and then add to the existing association set. For each new coming observation, we implement the clustering algorithm to update the associations. Each $\{ \cdot\}$ in the figure is one association set. }
	\label{fig:pipeline}
\end{figure*}

The operation of our algorithm is illustrated in Figure \ref{fig:pipeline}. Using the continuous input of an RGBD camera and 3D object detections from RGB and depth channels, a front-end module incrementally constructs a local object-level map. The latter is expressed in the form of a fully connected graph where nodes represent detected objects and edges encode relative spatial properties. Starting from an initial graph, object classes and shape descriptors are used in order to initialize association sets containing a single pairwise association between a detected object and a corresponding landmark in the global map. During hierarchical clustering, the relative spatial information between pairs of measured objects and pairs of corresponding objects in the global map is then used  to construct a merge metric and merge criterion for association sets. The hierarchical clustering is rerun when each time the camera measures a new set of objects, thus causing  the association sets to grow further and become more distinctive.

In the following, we will introduce the absolute and relative object information used during the initialization and clustering of the association sets, followed by the details of the hierarchical clustering itself. We conclude with details on the incremental update, the convergence criterion, and the actual pose recovery.

\subsection{Absolute and relative object properties}

\textbf{Absolute object position:}
In contrast to 2D feature point locations---which are potentially very accurate---the 3D position of an object is largely impacted by the partiality of the measurement. While we often set an object position $\mathbf{m}_i$ to the mean of an object's measured points, the characterisation of its uncertainty remains a challenging problem. This is especially true for large objects for which entire shape observations during normal camera navigation rarely occur. Absolute position uncertainty has an immediate impact on relative object properties such as the distance between two objects, which will be used later  to construct our merge metric. By assuming a maximum object shape size, this uncertainty can be bounded by a predefined threshold. Note that similar assumptions are common in related work. TEASER\cite{yang2020teaser} for example exploits a similar assumption in order to bound the influence of outliers during 3D-3D registration.

\textbf{Object descriptor:} We extract an object shape descriptor from each object point cloud $\mathbf{P}_i$, which maintains resilience against varying appearance information. Obtaining accurate object descriptors is substantially more difficult than point feature description and influenced by angles of observation, depth or segmentation noise, and occlusions. Traditional hand-crafted features cannot meet requirements in terms of expressiveness and stability. We typically employ deep neural networks for object shape encoding, such as DGCNN \cite{wang2019dynamic}, Pointnet++\cite{qi2017pointnet} and VoxNet\cite{maturana2015voxnet}. 
We choose VoxNet trained on the Scan2CAD dataset owing to its strong noise mitigation abilities. Representative and discriminative features---denoted $\mathbf{f}_i$---are obtained by using the triplet loss. The object class $s_i$ is used along with the object shape descriptor $\mathbf{f}_i$ in order to establish the initial, singleton association sets.

%
%
\textbf{Relative object position:}
We use object relative position information in order to express the objects' spatial distribution and construct a merge metric for our hierarchical clustering algorithm. We simply use the relative distance between two objects, which is invariant with respect to the camera pose or the frame of reference. For two objects $\mathbf{B}_i$ and  $\mathbf{B}_j$, the relative distance is simply given by $d^{B}_{i,j} = \|\mathbf{m}^B_i - \mathbf{m}^B_j \|$.
\subsection{Hierarchical clustering for association set growing}
\textbf{Initialization:}
Our purpose is to relocalize the camera, which we do by associating the current object observations with object instances in the global map.
We start by creating a set of hypothetical singleton association sets $\mathbf{D}^0=\{\mathbf{D}^0_k|k=1,\cdots, K_0\}$ by using object classes and shape descriptors to find individual matches for each detected object. For each potential match $(\alpha_i,\beta_i)$, the semantic labels need to be the same (i.e. $s_{\alpha_i}=s_{\beta_i}$), and the feature distance needs to be within a bound $\delta$ (i.e. $\|f_{\alpha_i}-f_{\beta_i}\|< \delta$). In order to handle situations of significant differences for example caused by severe partiality of an object's observation, $\delta$ is typically set to a relatively large value $0.5$.
The scores of each singleton association set will be initialized to $0$.  Note that---owing to potentially repeated occurrence of identical semantic labels---a single local instance may initially form multiple singleton association sets.

\textbf{Merge metric and merge criterion:}
After initialization, the algorithm will start to merge and generate new association sets by hierarchical clustering. The purpose of hierarchical clustering is to implicitly eliminate wrong associations by recursively finding pairs of spatially compatible association sets with maximum cardinality. All possible pairs of association sets will be recursively explored. The quality of an association set is determined by implementing a scoring mechanism alongside the clustering.
In classical clustering algorithms, one object node will only belong to one cluster center. In our implementation, the same association set could merge multiple times with other association sets if they satisfy the merge criterion.

The merge metric is formed as follows.
For any two potential merge candidates $\mathbf{D}_{k_1} = \{(\alpha_i,\beta_i)|_{i=1, \cdots, |\mathbf{D}_{k_1}|} ,v_{k_1}\}$ and $\mathbf{D}_{k_2} = \{(\alpha_j,\beta_j)|_{j=1, \cdots, |\mathbf{D}_{k_2}|} ,v_{k_2}\}$, we calculate the distance between each new pair of objects in both the local and the global map.
Note that new pairs of objects are defined by taking one element from each set such that neither the first element is contained in the second set, nor the second is contained in the first. We then take the distance ratio
\begin{equation}
h(\alpha_{i},\alpha_{j}, \beta_{i}, \beta_{j}) = min\left(d^{B}_{\beta_i,\beta_j}/d^{A}_{\alpha_i,\alpha_j},d^{A}_{\alpha_i,\alpha_j}/d^{B}_{\beta_i,\beta_j}\right),
\end{equation}
where $\alpha_i$ and $\alpha_j$ are the object indices in the global map, and $\beta_i$ and $\beta_j$ are the object indices in the local map. $d^{A}_{\alpha_i,\alpha_j}$ and $d^{B}_{\beta_i,\beta_j}$ are the object distances in the global and the local map, respectively. The merge metric expresses compatibility within a new pair of objects and has an upper bound of 1.
The $D_{k_1}$ and $D_{k_2}$ are merged if 
\begin{equation}
l(D_{k_1},D_{k_2}) == |D_{k_1}| * |D_{k_2}|,
\end{equation}
where
\begin{equation}
l(D_{k_1},D_{k_2})=\sum_{(\alpha_i, \beta_i) \in D_{k_1} (\alpha_j,\beta_j) \in D_{k_2}}   I (\gamma<h(\alpha_{i},\alpha_{j}, \beta_{i},\beta_{j})).
\end{equation}
$I (\gamma<h(\alpha_{i}, \alpha_{j}, \beta_{i},\beta_{j}))$ is an indicator function. If $\gamma<h(\alpha_{i},\alpha_{j} \beta_{i},\beta_{j})$, then $I(\cdot)=1$, else $I(\cdot)=0$.  In other words, association sets are only merged if they are both free of outliers and all new object pairs have a relative spatial arrangement that is consistent with their respective associated objects from the global map.
The score of the new association set is
\begin{equation}
v_{k_1} + v_{k_2} + \sum_{(\alpha_i, \beta_i) \in D_{k_1} (\alpha_j,\beta_j) \in D_{k_2}}   h(\alpha_{i},\alpha_{j},\beta_{i}, \beta_{j}).
\end{equation}

\textbf{Update:}
As the camera keeps exploring the environment, new object detections may occur. This will trigger a complete new iteration in which 1) the new or redetected instances will again be used to form singleton association sets that are added to the complete set of all association sets, and 2) another round of hierarchical clustering will be issued.

\textbf{Avoiding redundancy:} By default, the original association sets are not removed, even if they have been merged. This is because they could in principle form new interesting clusters once new object detections come in. However, in order to avoid redundant calculations, we mark clusters that have already been merged such that they will no longer be considered for merging during future updates. Note furthermore that one cluster may already contain another cluster, in which case no new object pairs can be formed, and no merging will be attempted. If a duplicate set is detected, we will manually remove the one with smaller score. We will furthermore remove sets that have not seen a single merge since more than two update steps. To conclude, we speed up execution by only taking the first top $K = 10*n$ association sets into the next round of clustering, where $n$ is the number of instances in the global map.  

\subsection{Camera pose recovery}
Note that global localization may be considered unambiguous and successful as soon as only a single association set with dominant score remains. We estimate the pose of the camera by running globally optimal ICP \cite{yang2015go} over the point-sets obtained by individually concatenating the point sets from the local and the global map for each corresponding object and then refine by  using robust ICP\cite{hu2021point}. 

%% file: articles/experiment.tex
\section{Experimental results}
\label{sec:experiment}
In the following, we briefly introduce the baseline methods that we compare against, the datasets on which we evaluate the relocalization, the evaluation metric, as well as all experimental results. 
\subsection{Baseline methods}
\subsubsection{Feature-based methods.}
For feature-based registration methods, we render several sub-maps from the global map. We sample points with grid step of $0.3$m in the global map and find their neighbors within a radius of $1.5$ meters to form one sub-map.
\begin{itemize}
\item \textbf{FPFH}~\cite{rusu2009fast} and \textbf{FCGF}~\cite{choy2019fully}: State-of-the-art hand-crafted and learned geometric  descriptors used for point cloud matching. We densely extract these descriptors and train a custom shape vocabulary on the global map. We also compare the traditional feature \textbf{SIFT} and use the corresponding depth points for later camera pose recovering.
\item \textbf{Semantic visual localization~\cite{schonberger2018semantic}:} For simplicity, we denote this work \textit{SVL}.
We reimplemented the method which  uses a variational auto-encoder to learn the representation of point observations. We randomly select 20 scenes from the ScanNet dataset and fuse several key-frames to obtain local semantic maps for training, and test its performance on other ScanNet scenes. 
\end{itemize}
Note that, in comparison to object-level relocalization, the above algorithms utilize the background information. As a result, they cannot be used if the pre-built map does not store this information or if the map is built by another sensor.

\subsubsection{Object-level relocalization.}
Object-level relocalization is done by using object features such as shape descriptors and---as done in our proposed method---relative object arrangements.
\begin{itemize}
    \item \textbf{SegMatch:}
SegMatch\cite{dube2016segmatch} is a place recognition algorithm based on the matching of 3D semantic segments. One-to-one correspondences are notably found by object classification. 
The code is obtained from an open-source framework and applied to indoor environments.
  \item \textbf{Random walk descriptors:}
Random walk descriptors make use of local topological graphs of object instances \cite{liu2019global}. 
The main shortcoming of the method is the requirement for a predefined walk length which is not suitable for the incremental relocalization.
We denote our re-implementation of this method as \textit{random-walk}, and use the original parameters of \cite{liu2019global}.
\end{itemize}
 \begin{figure}[htbp!]
	\begin{center}
    	\includegraphics[width=\columnwidth]{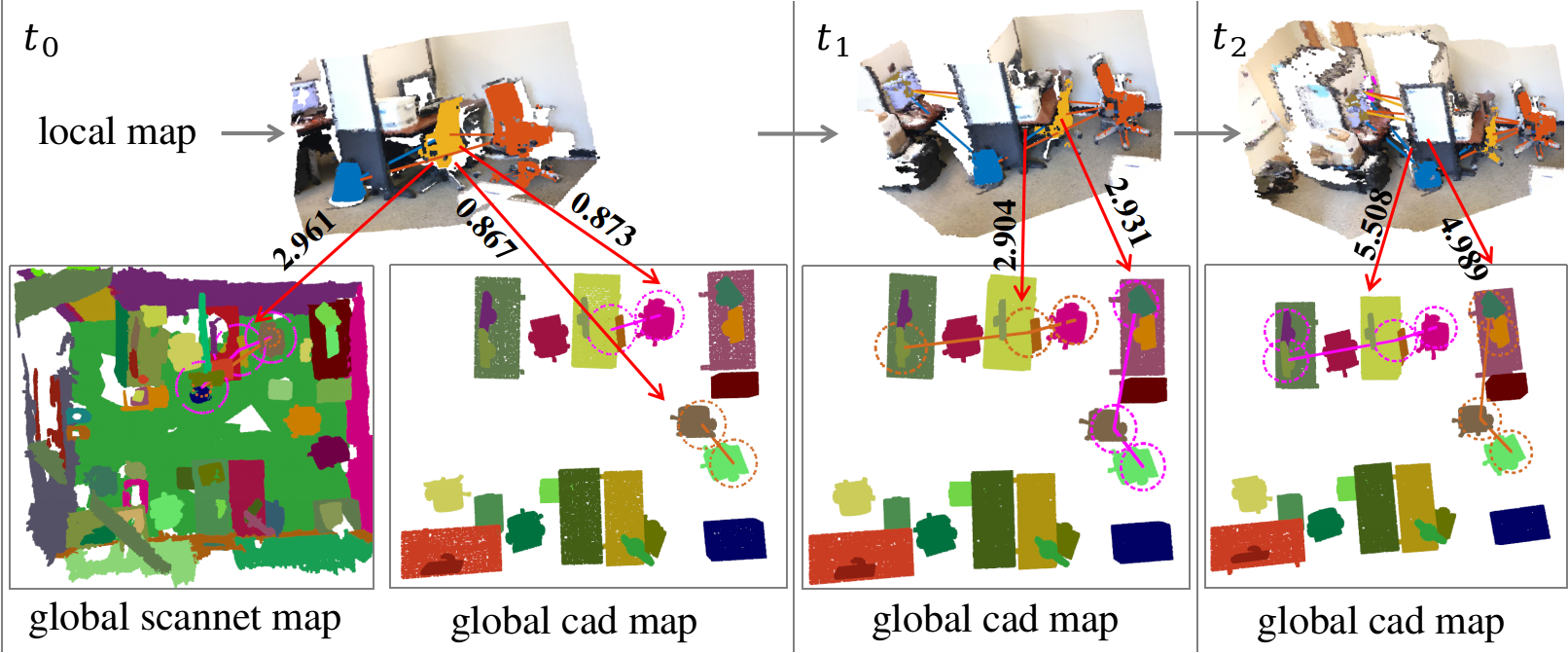}
	\end{center}
	\vspace{-0.2cm}
\caption{Visualization of clustering results on two different global maps. Owing to the existence of a unique semantic object in the global scannet map, using the latter leads to immediate convergence at $t_0$. For the CAD based map, the ambiguity needs further updates to be resolved.}
\label{fig:incremental_relocalization}
\vspace{-0.2cm}
\end{figure}
\subsection{Datasets and evaluation}
For the sake of a fair comparison, all algorithms use the same, previously outlined pose recovery technique. Common assumption with known ground plane\cite{hu2021globally,yang2018cubeslam,salas2013slam++} is made to alleviate the computation of camera pose estimation.
A successful retrieval is defined if the rotation and translation errors are below $15^{\circ}$ and $0.5$ meters, respectively.
We test our algorithm on the ScanNet\cite{dai2017scannet} and SSS\cite{cao2020representations} datasets. ScanNet is a dataset with real images observing various indoor scenes. SSS provides indoor scenes in day and night situations which are rendered by \textit{Blender}. Both SSS and ScanNet have two global maps: a geometrical object-level reconstruction, and a semantic model map in which the objects are replaced by their corresponding CAD models. The ground truth CAD models of ScanNet are taken from the Scan2CAD~\cite{avetisyan2019scan2cad} dataset. Models of the SSS dataset are exported from \textit{Blender}.

It is intuitively clear that owing to the robustness of instance-level segmentation, object-level relocalization is much less affected by view-point or illumination changes than key-point feature-based relocalization. We validated this statement on the SSS dataset \textit{room5}, a relatively large indoor scene that extends over two rooms. We rendered both \textit{day} and \textit{night} sequences and defined orbiting camera trajectories in each room that capture an abundance of views of an identical place but under very different viewing angles. Through our tests, we were able to verify that instance-level segmentation results are much more robust than feature point detections.

\subsection{Front-end}
We use BlendMask~\cite{chen2020blendmask} to provide instance-level semantic segmentation from RGBD images. Note however that RGB information is not used during any later stages of our relocalization pipeline, which is why the use of other instance-level segmentation networks suitable for pure depth data~\cite{jiang2020pointgroup} could easily extend our approach to scenarios in which no regular images are being captured. Our local instance-level map is built using an approach inspired by the works of Grinwald et al. \cite{grinvald2019volumetric} and Hu et al. \cite{hu2019deep}.
While instance-level segmentation frameworks generally have good robustness against illumination or view-point variations, they still pose challenges in the form of occasionally unstable performance, wrong or even missing detections, and bad segmentation masks. 
The influence of the segmentation quality on semantic relocalization is well documented in Liu et al. ~\cite{liu2019global}.

\subsection{Visualizing the effect of hierarchical clustering}
Our hierarchical clustering algorithm grows and scores association sets. Its inherent functionality makes it suitable to gradually account for the results of incremental local mapping, and thereby identify an association set with high confidence. To confirm its ability to gradually localize the camera with higher confidence,  let us see the scores of the top two association sets.  Figure \ref{fig:incremental_relocalization} visualizes incremental association sets on ScanNet scene \textit{scene0475\_00}, which contains multiple instances of the same object.  

As indicated in Figure \ref{fig:incremental_relocalization}, the scores for the possible association set will become increasingly discriminative as more objects are being observed.
In particular, the correct association set will be quickly identified if the camera observes an object that is unique in one of the maps.
\begin{figure}[htbp!]
	\begin{center}
    	\includegraphics[width=0.85\linewidth]{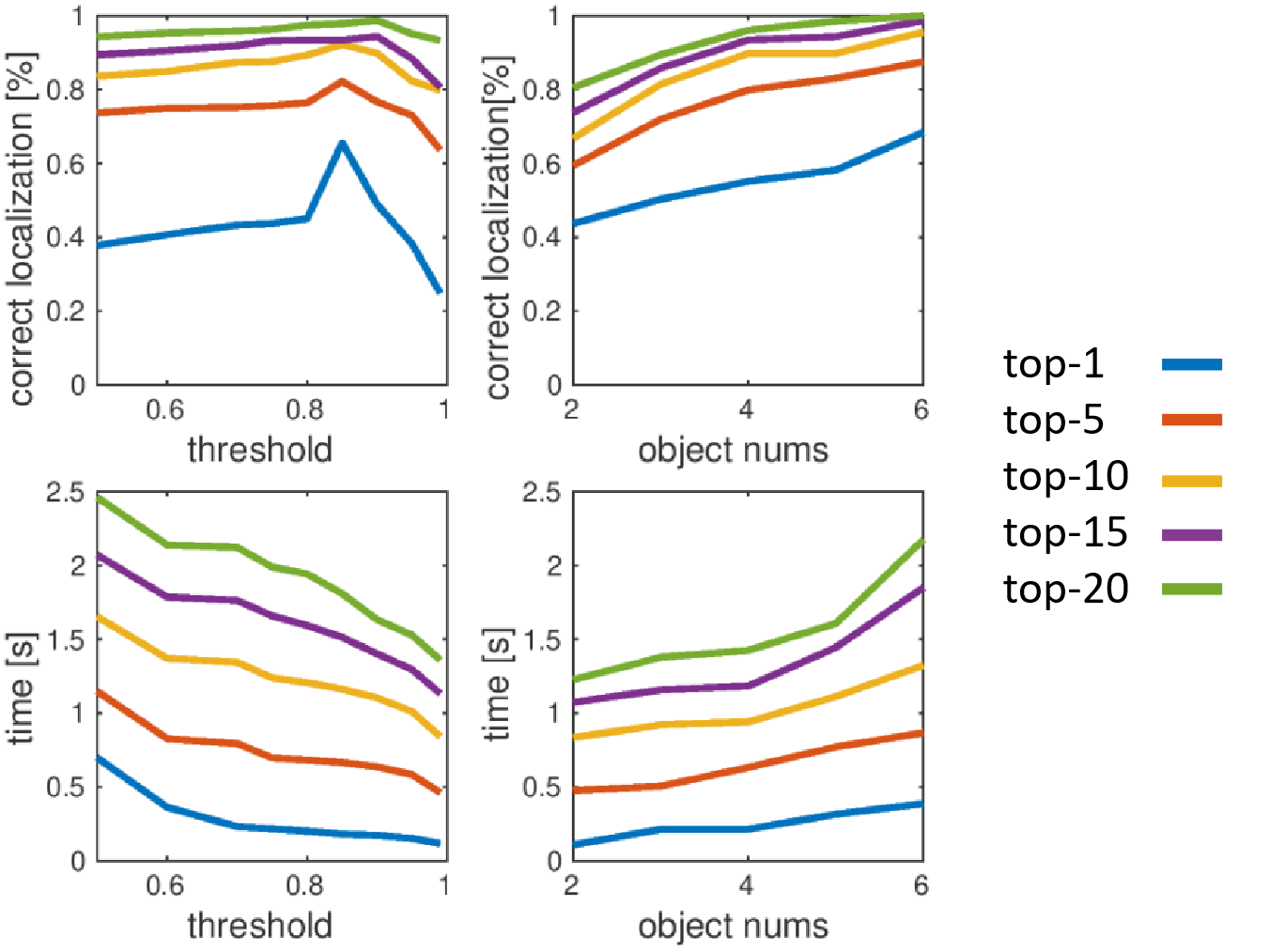}
	\end{center}
\caption{Ablation study. Here we analysis the performance influence from clustering threshold, the object number in the local map and their corresponding time costing.}
\label{fig:analysis}
\end{figure}
\subsection{Ablation study}
Both the number and the position uncertainty of objects will impact on the relocalization quality. Results are further depending on the chosen value for $\gamma$. We therefore perform an ablation study over $\gamma$ and the number of objects on ScanNet \textit{scene0247\_01} and \textit{scene0365\_01}. 

Success is given if one of the choosen top-k association sets  for camera pose recovering is the correct location, where top-k $ \in [1,5,10,15,20]$. The larger top-k will bring higher success rate.
Figure \ref{fig:analysis} shows the experiment results. The first row indicates the relocalization success ratio, while the second row analyses the computational cost. 
Choosing only the top-1 association set for pose recovering will obviously lead to the lowest success ratio, whereas as soon as top-k $ \geq 10$, results become very identical.
As the left column shows, a tighter $\gamma$-threshold will generally lead to higher localization success ratio, thought if the threshold is too tight, the success ratio will again decrease. The threshold should be set smaller than 0.9, and---to take more potential associations into account---it is set to $0.8$ in all remaining experiments.
Note that the indicated time cost contains the time for both the hierarchical clustering and pose recovery. The time depends on the number of considered association sets, and is lowest if top-k $=1$. It is furthermore obvious that as the threshold $\gamma$ becomes tighter, the time cost will decrease as fewer association sets are generated. The right column of  Figure \ref{fig:analysis}  presents the analysis on how the number of objects influences the success rate and the computational cost. More objects will clearly increase the relocalization success rate at the cost of higher computational demands.
\begin{figure}[htbp!]
	\begin{center}
    	\includegraphics[width=\columnwidth]{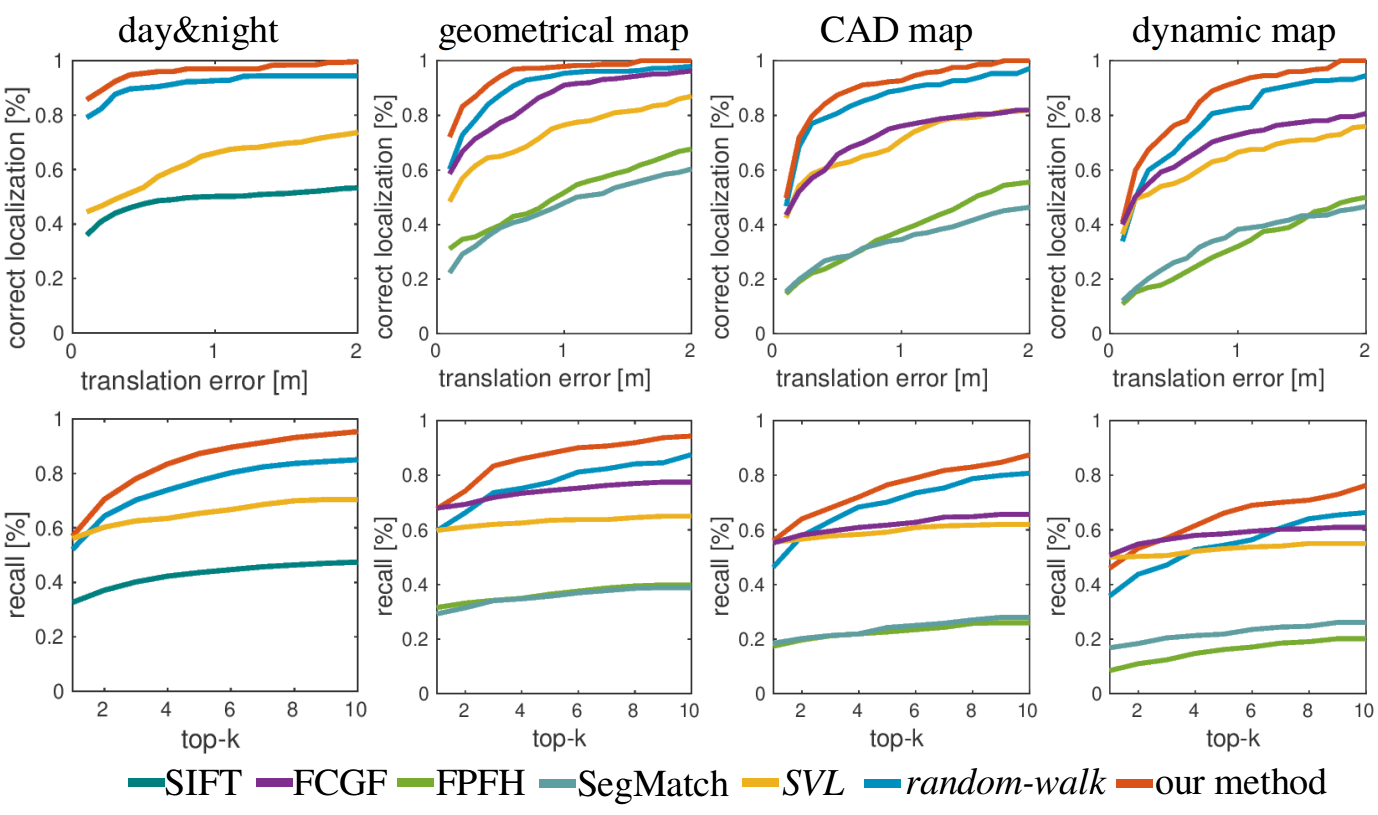}
	\end{center}
\caption{Comparison against alternative methods.}
\label{fig:comparision}
\end{figure}
\subsection{Comparison against alternative methods}
To conclude, we compare our algorithm against all baseline algorithms on several datasets including different illumination conditions and three different types of global map: the geometrically reconstructed map, the cad model map, and an extended cad model map in which objects have been moved (denoted dynamic map). To obtain the dynamic map, we take two steps. First, we delete some objects from the global map. Second, we randomly move some of the remaining objects to new positions. We make sure that at least $30\%$ of the objects are moved to new places. A total of $7$ different sequences with multiple same objects are analyzed, which are \textit{room5 day} and \textit{room5 night} for our day-night experiments, and  $5$ ScanNet sequences
(\textit{scene0475\_00}, \textit{scene0365\_01}, \textit{scene0247\_01}, \textit{scene0603\_01} and \textit{scene0250\_01})
for all other experiments. Results are averaged over more than $3000$ relocalization attempts. Relocalization attempts are started from uniform randomly picked key-frames along each sequence. For each starting point, we will generate three relocalization attempts. The first one happens as soon as more than $2$ objects are observed. The second and third localization attempts happen by taking an additional $15$ or $30$ key-frames into account.

Figure \ref{fig:comparision} shows the average results for the correct localization rate or recall over the tolerated translation error (row 1, top-$k$=10), as well as the considered number $k$ of top-ranked association sets (row 2, error tolerance: $0.5$ m and $15^{\circ}$).
The first column shows the relocalization result during night where the global map has been reconstructed during the day. As can be observed, our method performs very well while traditional point-feature based methods such as SIFT are unable to deal with the variation. Here we did not present the result of 3D feature-based methods, but we show in the second column, because they will not be affected by the illumination changes.
The next three columns are the results referring to the different types of global map.  Overall, we can observe that our object-level relocalization has better performance:
\begin{itemize}
    \item As expected, the hand-designed FPFH feature performs worse. The 3D point map contains many repeated, similar local structures, which causes the  hand-designed  feature FPFH to be not representative enough for relocalization. FPFH perform even worse when the global map is represented by a CAD model, which typically leads to differing point distributions in the locally observed map.
    \item Our experimental results indicate that \textit{SegMatch} is troubled by multiple similar instances and the lack of shape differences, since it extracts several features for each instance and then uses a random forest to find an instance association between the local observation and the global map. Such strategy is easily influenced by the camera view and the object class, where different camera views will result in feature  dissimilarities, and multiple same-class objects hold the ability to confuse the algorithm and cause the wrong final association decisions.
    \item Learning-based methods, such as \textit{SVL} and FCGF, have very close performance and FCGF obtains better performance than our algorithm when top-k is less than $4$ (cf. fourth column of Figure \ref{fig:comparision}, which occurs in the dynamic situation). This may be because \textit{SVL} and FCGF rely on a lot of background information rather than just object information. However, compared against our light-weight algorithm, the learning-based alternative needs larger computation resources and will thus be limited by the platform. 
    \item The most related work to ours is \textit{random-walk}, which also exploits object-related spatial distributions. However, it depends on a predefined walk length and does not score each association set. Our algorithm on the other hand adopts an incremental clustering strategy, and scores all association sets. The more observations, the more certain the correct association will be.
\end{itemize}

The dynamic situation is exposed in the fourth column. Most algorithms perform worse in this scenario. However, our algorithm still provides competitive performance especially when the local observed object number is larger.

\section{Conclusion}
We present a novel global localization framework that is conceptually simple and relies on object detections and associations with objects in a global map. The overall success rate of the method is high and it easily outperforms existing methods as soon as sufficiently many objects can be detected, irrespectively of their class. The method operates very fast and is particularly well suited in conjunction with incremental, object-level mapping front-ends. We believe it could be of broad interest in object-level SLAM frameworks.